\documentclass{article}

 \PassOptionsToPackage{numbers, compress}{natbib}

\usepackage[preprint]{nips_2018}

\usepackage[utf8]{inputenc} 
\usepackage[T1]{fontenc}    
\usepackage{hyperref}       
\usepackage{url}            
\usepackage{booktabs}       
\usepackage{amsfonts}       
\usepackage{nicefrac}       
\usepackage{microtype}      

\title{Clustering by latent dimensions}

\author{
Shohei Hidaka\thanks{ Use footnote for providing further information
about author (webpage, alternative address)---\emph{not} for acknowledging
funding agencies.} \\
School of Knowledge Science\\
Japan Advanced Institute of Science and Technology\\
1-1 Asahidai, Nomi, Ishikawa, Japan \\
\texttt{shhidaka@jaist.ac.jp}
\And
Neeraj Kashyap \\
Doc.ai \\
636 Waverly Street, Suite 200 \\
Palo Alto, CA 94301 \\
\texttt{neeraj@doc.ai}
}

\usepackage[dvipdfmx]{graphicx}
\usepackage{amssymb,amsfonts,amsmath,amsthm}
\usepackage{mathrsfs}
\usepackage{array}

\def\diam{\mathrm{diam}}

\newcommand{\RNDim}{K}
\newcommand{\RNN}{\mathbb{R}^{\RNDim}}
\newcommand{\RN}{\RNN}

\def\Dim{d}
\def\Num{N}
\def\dimclust{dimensional clustering{}}

\newtheorem{assumption}[]{Assumption}

\begin{document}

\maketitle

\begin{abstract}
This paper introduces a new clustering technique, called {\em dimensional clustering}, which clusters each data point by its latent {\em pointwise dimension}, which is a measure of the dimensionality of the data set local to that point. Pointwise dimension is invariant under a broad class of transformations. As a result, dimensional clustering can be usefully applied to a wide range of datasets.
Concretely, we present a statistical model which estimates the pointwise dimension of a dataset around the points in that dataset using the distance of each point from its $n^{\text{th}}$ nearest neighbor.
We demonstrate the applicability of our technique to the analysis of dynamical systems, images, and complex human movements.
\end{abstract}

\section{Introduction}

The general idea in statistical modeling is to:

\begin{enumerate}

\item{ Identify a spatial representation of your dataset which is suitable for analysis}

\item{ Identify a class of probability distributions over your represenation space that could
be considered descriptive of the dataset}

\item{ By some procedure of your choosing, estimate which probability distributions in the chosen
class are most likely to be the distribution from which the dataset was sampled}

\end{enumerate}

Domain knowledge has traditionally been very important in steps 1 and 2 above -- the crucial choices
of the data representation and the family of admissible probability distributions. The human analyst
has had a central role in these decisions.

However, cutting edge machine learning models like the one employed by AlphaGoZero \cite{Silver2017}
attempt to surpass the knowledge of human experts. In the case of AlphaGoZero, it has learned how to
play the game of Go without observing any games in which humans were involved. It only played
itself.

This study follows this vein of ideas. We develop a technique of statistical modeling, \dimclust,
which:
\begin{enumerate}

\item{ Has minimal dependence on the representation of the data or the family of descriptive
probability distributions.}

\item{ Can be applied to a wide range of datasets.}

\end{enumerate}

Such a technique, by nature, must rely on as small a set of assumptions regarding the dataset as
possible. Our technique, \dimclust, makes only the following assumption of the dataset that is the
subject of its analysis:

\begin{assumption}
The dataset consists of points in some metric space $X$ that were sampled\footnote{ This also
includes points in an orbit of an ergodic dynamical system on $X$.} from some Borel measure on $X$.
\end{assumption}

\dimclust {} is a clustering technique, and clusters points in the dataset by a local fractal dimension,
called \emph{pointwise dimension}. Pointwise dimension is invariant under bi-Lipschitz transformations
between metric spaces and under topologically equivalent choices of metric on
a given space. This means, for example, that the pointwise dimension around a given point will be
the same as the pointwise dimensions around its images under continuous deformations of the dataset.

The following section describes more specifically how we estimate pointwise dimension, but it is worth
pointing out here that this property of pointwise dimension being invariant over all choices of
topologically equivalent metrics means that, unlike traditional clustering techniques like $k$-means
clustering, \dimclust\ depends very little on the metric one imposes on the dataset.

Pointwise dimension is not the only fractal dimension that can be used in data analysis. There are
other quantities, such as \emph{correlation dimension} \cite{Grassberger1985}, which also reflect
the dimensionality of the dataset as a whole. However, none of them are as local as pointwise
dimension, and are therefore more suceptible inaccuracies due to noise in the dataset \cite{Kantz1993}.
Because pointwise dimension is a local property, it can be derived from local statistics of the
dataset. The derivation we present here is based on nearest neighbor distances.

Our paper is laid out as follows:
In Section \ref{sec-Overview}, we provide a brief overview of the theory behind pointwise dimension and derive the statistical model for pointwise dimensions based on nearest neighbor
distances.
In Section \ref{demo}, we demonstrate how pointwise dimension is estimated over a dataset sampled from the Cantor measure.
In Section \ref{sec-Algorithm}, we describe an algorithm implementing dimensional clustering
in some detail. The algorithm has virtually no paramaters that one has to manually tune for each
dataset. It automatically estimates the number of clusters that should be used and the probability
of assignment of each point in the dataset to each cluster.
In Section \ref{sec-Application}, we apply \dimclust {} to multuple datasets.

\section{Pointwise dimension}
\label{sec-Overview}

Fractal dimensions, like the Hausdorff dimension \cite{Hausdorff1918}, where conceived as invariants that one could use to study sets, 
like the Cantor set,  which did not yield to useful measure-theoretic descriptions. We will use the Cantor set as an example illustrating 
the calculation of Hausdorff and pointwise dimensions.


The derivation of the Hausdorff dimension of the Cantor set goes as follows. The Cantor set (which will be described with more precision 
later) is too ``sparse'' to consider a 1-dimensional object.
However, it is also too ``dense'' (in the sense that it consists of uncountably many points) to count it as a 0-dimensional object.
Informally speaking, that means the Cantor set is an object of dimensionality higher than zero and lower than one.
Hausdorff's idea was to evaluate such sets using the continuous family of measures which interpolate between these concepts of
0-dimensional size (cardinality) and 1-dimensional size (length). These measures, called Hausdorff measures \cite{Hausdorff1918} 
$H^{\alpha}$, are defined as follows:

Let $U \subseteq \RN$, and let $\epsilon > 0$. An {\em $\epsilon$-covering} of $U$ is a countable collection $\{U_{k}\}_{k}$ of sets that $U \subseteq \cup_{k}U_{k}$ and $\sum_{k}\diam(U_{k}) \le \epsilon$, where $\diam( U ) = \sup_{x, y \in U} \|x - y\|$ with some metric $\| \circ \|$. For $\epsilon > 0$, put
\[
 H^{\alpha}( U ) := \lim_{\epsilon \to 0}\inf_{\{U_{k}\}}
   \left\{
     \sum_{k}\diam(U_{k})^{\alpha} \mid \{U_{k}\}_{k}
   \right\}
\]
One finds the Hausdorff measure $H^{\alpha}( U )$ of any set $U$
is a non-negative, monotone decreasing function of $\alpha$. Crucially, $H^\alpha$ can be finite positive with at most one
$\alpha \in [0, \infty)$.
This critical value $\hat{\alpha} = \sup\{ \alpha : H_\alpha = 0 \}$ is called Hausdorff dimension, which gives a natural extension of ``dimension'' notion $\RNDim$ of $\RN$.
For the Cantor set $U$, we have finite positive measure $H(U)^{\hat{\alpha}}$ with $\hat{ \alpha } = \log(2)/\log(3)$.

The limit over coverings of $U$ in the definition of Hausdorff measures makes Hausdorff dimension very difficult to calculate directly or
numerically. To mitigate this difficulty, several substitutes for Hausdorff dimension have been introduced, especially in the literature
surrounding non-linear dynamical systems. Premier amongst these is correlation dimension, introduced by Grassberger and Procaccia 
\cite{Grassberger1985}.
In this study, we consider a different quantity, {\em pointwise dimension}, defined as follows \cite{Cutler1993,Young1983}.


Let $\mu$ be a Borel probability measure on $\RN$. 
For $x \in \RN$, let $B(x,\epsilon)$ denote the $\epsilon$-ball around $x$. The pointwise dimension of $\mu$ at $x$ is defined, if it exists, as the 
limit
\begin{equation}
\label{PointwiseDimension}
\Dim_{\mu}(x) := \lim_{\epsilon \rightarrow 0} \frac{\log\mu\left(B(x, \epsilon)\right)}{\log \epsilon}.
\end{equation}

Pointwise dimension $\Dim_{\mu}(x)$ is a local quantity which describes the scaling behaviour of the measure $\mu$ at a point $x$.
The pointwise dimension \emph{at any point} $x$ in the Cantor set is $\log(2)/ \log(3)$, which agrees with its Hausdorff dimension.

This description, by itself, does not solve the problem of having to resolve a limit at infinitesimal scale -- such limits are unavoidable
when defining any fractal dimension. Dealing with such limits is particularly difficult when working with real-world datasets as even
small amounts of noise can result in gross misestimations of the limiting behaviour.
However, pointwise dimension can be used to overcome the difficulty of numerical calculation of fractal dimension by means of the
statistical estimator devised in Hidaka and Kashyap \cite{Hidaka2013}. In that work, the measure $\mu$ is assumed to be a convex
combination of locally uniform measures, which have constant pointwise dimension over their supports.
For each of these locally uniform components, the limiting problem is solved by estimating the pointwise dimension from the nearest 
neighbor distance statistics of the dataset instead of an extrapolation of its infinitesimal limiting behaviour.

Given a locally uniform measure of pointwise dimension $\Dim$ over its support, a point $x$ in its support then there exists $\rho > 0$ 
such that, for sufficiently small $r > 0$, $p(r) := \rho r^{\Dim}$ is a sufficiently good estimate for the measure 
of the $r$-ball around $x$.
For a sufficiently large number $k$ of samples within a small distance $\epsilon$ of $x$, the cumulative distribution function for the 
distance of the $n^{\text{th}}$ nearest neighbor in the sample to $x$ is approximately
\begin{eqnarray}
F_{n}(r) = \sum_{m = n}^{k}{k \choose m} \left( 1 - p(r) \right)^{k-m}p(r)^{m}
.
\end{eqnarray}
%
Fixing $\lambda := (k-n)\rho$ and taking the number of samples $k$ to be increasingly large, the derivative of the limit of $F_n(r)$ with 
respect to $r$ give the probability density function
\begin{eqnarray}
P(r \mid n, \Dim, \lambda ) := \frac{\partial}{\partial r}
\lim_{k\rightarrow \infty}F_{n}(r) =
\frac{ \Dim \lambda^{n} \epsilon^{n\Dim - 1} }{ (n-1)! } e^{ - \lambda \epsilon^{ \Dim } }.
\label{eq-nndist}
\end{eqnarray}

Given a dataset $r_{1}, \ldots, r_{k}$ of the $n^{\text{th}}$ nearest neighbor distances for a set of points, this gives us a statistical 
model that we can use to estimate $d$ and $\lambda$.
For the special case $n=1$ (the first nearest neighbor distance), which we use for the data analyses in the following sections,
this probability distribution (\ref{eq-nndist}) is reduced to a Weibull distribution.

\section{Demonstration: Cantor measure \label{demo}}



In this section, we derive an estimate for the pointwise dimension of the Cantor middle-thirds set around each of its points. We do so
using the nearest neighbor distance (\ref{eq-nndist}) with $n = 1$.

First, we recall the construction of the Cantor set. The Cantor middle third set is a set of all the real numbers in the interval $[0,1]$, 
which have ternary representations in which none of the digits are 1. The Cantor set can be constructed iteratively as follows:

Put $I_{1}^{(0)} := [0,1]$. Define $M_{0}( [a, b] ) := [1/3a + 2/3b, b]$ and $M_{1}( [a, b] ) := [a, 2/3a + 1/3b]$.
Repeatedly take $I^{(i+1)}_{2j-1} = M_{0}( I^{(i)}_{j} )$ and $I^{(i+1)}_{2j} = M_{1}( I^{(i)}_{j} )$ for every $1\le j \le 2^{i}$.
The Cantor set is then given by the intersection $\cap_{i=0}^{\infty}\cup_{1\le j \le 2^{i}}I^{(i)}_{j}$.

The sums of lengths the intervals $I^{(i)}_j$ at the $i^{\text{th}}$ step of this construction is $(2/3)^{i}$, which approaches 0 as
$i \to \infty$, proving that the Cantor set has Lebesgue measure 0.

The number of points in the Cantor set corresponds to the number of choices of each of the infinitely many ternary digits that a number
in the unit interval can have and, even up to the countable number of reductions involving infinite tails of 2's, this is in bijection
with the unit interval. The Cantor set is therefore uncountable.

This is the sense in which the Cantor set is too ``sparse'' to be considered 1-dimensional, and too ``dense'' to be considered
0-dimensional.

Suppose that we sample ${\Num}$ points from the Cantor set. Using the iterative construction of the Cantor set, we see that
the probability that at least one of the points in the sample lies in the same interval $I_j^{(i)}$ as any given point $x$ is
\[ \mathrm{Pr}( r \leq 3^{-i} ) = 1 - (1 - 2^{-i})^{\Num}.\]
Thus, the probability mass of the nearest neighbor distance $r$ is
\begin{equation}
\label{eq-probmass}
\mathrm{Pr}( 3^{-i-1} < r \leq 3^{-i} ) = (1 - 2^{-i-1})^{\Num} - (1 - 2^{-i})^{\Num}.
\end{equation}

\begin{figure}[htb]
   \begin{center}
    \includegraphics[width=0.6\linewidth]{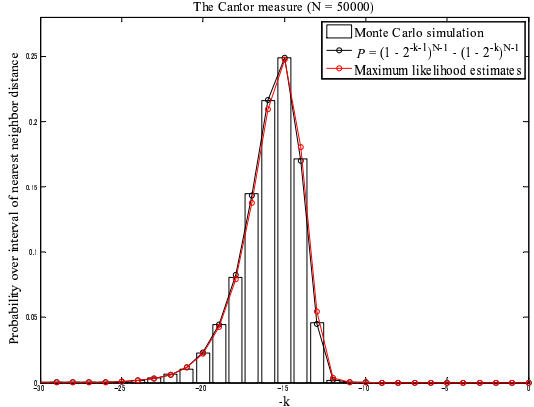}
    \caption{ The probability distribution of the nearest neighbor distance of the Cantor set. \label{fig-CantorExample}}
    \end{center}
  \end{figure}

Figure \ref{fig-CantorExample} shows this theoretical probability distribution (\ref{eq-probmass})
for a sample of ${\Num} = 50,000$ points randomly drawn from the Cantor measure (black circles) and the histogram constructed by an equivalent Monte Carlo simulation.
For the sample generated by the Monte Carlo simulation, we computed the nearest neighbor distance $r_{i}$ ($n=1$) for each point.
We used these distances to estimate $d$ and $\lambda$ by maximizing the log-likelihood function
$\sum_{i=1}^{50,000} \log P(r_{i} | n, d, \lambda )$ using the approximation (\ref{eq-nndist}).
This produced the estimates $\hat{d} = 0.62422$ and $\hat{\lambda} = 2.5567 \times 10^{-7}$ for the dimension and scale respectively.
Note that the estimated dimension $d = 0.62422$ is close to the true value $\frac{\log(2)}{\log(3)} \approx 0.6309$ for the
pointwise dimension of the Cantor set.

We used the estimates $\hat{d}$ and $\hat{\lambda}$ in (\ref{eq-nndist}) to compute an approximation to the probability mass function of 
the nearest neighbor distance $r$. This approximation is denoted by the red circles in Figure \ref{fig-CantorExample}.
Note that this estimated probability mass function closely fits with the theoretical one underlying the sample (Eq (\ref{eq-probmass})).

In order to apply this idea to a more general class of datasets, we must extend the model to capture multiple distinct locally uniform measures. This motivates the development of the mixture model introduced in the next section.

\section{Maximum likelihood estimator of pointwise dimension}
\label{sec-Algorithm}

\subsection{Statistical model}

A single dataset may consist of data generated by multiple locally uniform components, each with distinct 
parameters $d$ and $\lambda$. Note that there is a natural hierarchy in in these distinctions, in the sense that 
the scale parameters $\lambda$ are only comparable across locally uniform measures which have the same pointwise 
dimension. Our model is designed to capture this hierarchical relationship between $d$ and $\lambda$. 
The model consists of $M$ probability distributions, each having a probability density function
of the form (\ref{eq-nndist}). For each component, we index the scale parameters as $\lambda_l$ for $1 \leq l \leq M$.
However, the dimension parameters are indexed as $d_{c}$, where $h$ is a surjective map $l \in \{1, \ldots, M\} \mapsto c \in \{1, \ldots, M_d\}$, $M_d \leq M$ is the number of different dimensional parameters which appear amongst the components, and $c = h(l)$.

We decompose the probability density function for the $n^{\textrm{th}}$ nearest neighbor with respect to a large
sample from the generating measure as
\begin{equation}
\label{eq-statistical-model}
 P(r_{i} \mid n, (\lambda_{l}), (d_{h(l)}), (\theta_{l}) ) = \sum_{l = 1}^{M}\theta_{l} P(r_{i} \mid n, d_{h(l)}, \lambda_{l} ),
\end{equation}
where the $\theta_{l}$ are non-negative weight parameters with $\sum_{l}\theta_{l} = 1$.

Using (\ref{eq-statistical-model}), the log-likelihood function of the the parameters for a list of the $k$ $n$-th nearest neighbor distances $r_{1}, \hdots, r_{k}$ is
\begin{equation}
\label{eq-loglike}
 L( (d_{h(l)}), (\lambda_{l}), (\theta_{l}) ) = \sum_{i=1}^{k}\log P(r_{i} \mid n, (d_{h(l)}), (\lambda_{l}), (\theta_{l}) ).
\end{equation}


\subsection{Parameter search}
In this study, we maximize this likelihood function by the Expectation-Maximization (EM) algorithm \cite{Dempster1977}.
Introducing the non-negative conditional weights $Q_{li}$ ($\sum_{l}Q_{li} = 1$),
apply the Jensen's inequality and have the lower bound $\hat{L}( (\lambda_{l}), (d_{c}), (\theta_{l}), (Q_{li}) )$
for the log-likelihood function
\[
 L( (d_{h(l)}), (\lambda_{l}), (\theta_{l}) ) \ge \sum_{i=1}^{k} \sum_{l = 1}^{M}
\left\{
 Q_{li}\log \theta_{l}P(r_{i} \mid n, \lambda_{l}, d_{h(l)} ) - Q_{li}\log Q_{li}
\right\}
 =: \hat{L}.
\]
In the Maximization (M) step, we maximize $\hat{L}$ with respect to $(d_{h(l)}), (\lambda_{l}), (\theta_{l}) $ given $( Q_{li} )$.
In the Expectation (E) step, we maximize it with respect to $(\lambda_{l}), (d_{c})$ given the others.

At the initial step, let $( Q_{li} )$ be randomly generated values.
Then, given $( Q_{li} )$ in the M step, update the parameters by
\[
 \hat{ \theta }_{l} = \frac{ \sum_{i=1}^{k} Q_{li} }{ \sum_{l=1}^{M}\sum_{i=1}^{k} Q_{li}},
\hat{ \lambda }_{l}
= \frac{ n \sum_{i=1}^{k} Q_{li} }{ \sum_{i=1}^{k} Q_{li} r_{i}^{d_{h(l)}} }.
\]
For the dimension parameter $d_{h(l)}$, there is no closed form for the maximizer unlike $\hat{\theta}_{l}$ and $\hat{\lambda}_{l}$.
Thus we compute the maximizer by applying Newton's method to $\frac{\partial \hat{L}( ( d_{c} ), ( \hat{\lambda}_{l} ), ( \hat{\theta}_{l} ) )}{\partial \log d_{c}} = 0$ for each $c$.
Specifically, let $d_{c}^{(0)}$ be some positive constant, and
the maximum likelihood estimator $\hat{d}_{c} = d_{c}^{(\infty)}$ is given by the series $t = 0, 1, \hdots$
\[
\log d_{c}^{(t+1)} = \log d_{c}^{(t)}
+
\frac{
 \sum_{i=1}^{k}
\sum_{l \in \{l \mid h(l)=c\} } Q_{li}
\left(
  d_{c} \log r_{i}
-  d_{c} E[ \log x ]
 +
 n^{-1}
\right)
}
{
\sum_{i=1}^{k}
\sum_{l \in \{l \mid h(l)=c\} } Q_{li}
\left(
  d_{c}^{2} E[ ( \log x )^{2} ]
-  d_{c} E[ \log x ]^{2}
%
+ n^{-1}
\right)
}
,
\]
where $E[ f(x) ] = \frac{ \sum_{i=1}^{k} Q_{li} r_{i}^{d_{c}} f( r_{i} ) }{ \sum_{i=1}^{k} Q_{li} r_{i}^{d_{c}} }$ with an arbitrary function $f(x)$.

In the E step, given $\hat{d}_{h(l)}, \hat{\lambda}_{l}, \hat{ \theta }_{l}$, update $Q_{li}$ for each $i = 1, \ldots, k$ and for each $l = 1, \ldots, M$ by
\[
\hat{Q}_{li} \propto
  \hat{ \theta }_{l} P(r_{i} \mid n, \hat{\lambda}_{l}, \hat{d}_{h(l)} ) .
\]
Repeat the M step and E step above until the difference in the log-likelihood $\hat{L}(t) - \hat{L}(t-1)$ of step $t$ and $t-1$ is sufficiently small. In this study, the iteration is stopped if $\hat{L}(t) - \hat{L}(t-1) \le 10^{-5}$ or $t \ge 5000$.

For a given pair of numbers of clusters $(M_{d}, M)$, the maximal log-likelihood is given by $L((\hat{d}_{h(l)}), (\hat{\lambda}_{l}), (\hat{\theta}_{l}))$ (\ref{eq-loglike}). 
The dataset is decomposed into $M_d$ dimensional clusters with the $c^{\text{th}}$ dimensional cluster
($1 \leq c \leq M_d)$ being defined by the components in (\ref{eq-statistical-model}). For the $c^{\textrm{th}}$
dimensional cluster, the dimension is $d_{c}$, rates are $\lambda_l$, and weights are $\theta_{l}$,
for all $1 \leq l \leq M$ such that $h(l) = c$.
The probability of the $i^{\textrm{th}}$ point belonging to the $c^{\textrm{th}}$ dimensional cluster is given
by $\sum_{l \in \{ l \mid h(l) = c \}} \hat{Q}_{li}$.
The average pointwise dimension over the entire dataset is defined by $\sum_{i=1}^{k}\sum_{l=1}^{M}Q_{il}d_{h(l)}$.

\subsection{Model search}

Using the maximal log-likelihoods (\ref{eq-loglike}), we can also search the space of possible model structures
given by the pairs $(M_d, M)$. We do this by searching for the optimal $M_d$-tuple $(m_1, m_2, \ldots, m_{M_d})$
where $m_1 \geq m_2 \geq \ldots \geq m_{M_d}$ and $\sum_{i=1}^{M_d}m_i = M$. These integers $m_i$ represent
the numer of scale components in each dimensional cluster.

We perform our search by minimizing the AIC \cite{Akaike1974} given by
\[\text{AIC}(m_{1}, m_{2}, \ldots, m_{M_{d}}) = -2 L((\hat{d}_{h(l)}), (\hat{\lambda}_{l}), (\hat{\theta}_{l})) + 2( M_{d} + 2M - 1 ).\]

Specifically, we initialize our search with a model having the structure $M_{d} = 1$ and $(m_1) = (1)$. Assuming a
model structure of $(m_1, m_2, \ldots, m_{M_d})$ at step $t$ of the model search, we evaluate the AICs of all
possible models with structure $(n_1, n_2, \ldots, n_{M_d+1})$ subject to the constraints
$n_1 \geq n_2 \geq \ldots \geq n_{M_d+1}$, $n_i \geq m_i$ for $1 \leq i \leq M_d$, and $\sum_{i=1}^{M_d+1}n_i = M+1$. If none of these model structures yields a model with lower AIC than the minimal AIC model at the current
step $t$, we stop the model search and choose the current model as the best one. Otherwise, we let the model
structure chosen at step $t+1$ of the model search be the minimal AIC model over the new structures. This effects
a breadth-first search over the implicit graph of model structures with edges defined by the constraints on the
tuple $(n_1, n_2, \ldots, n_{M_d+1})$. It terminates at a structure with locally minimal AIC.

\section{Application}
\label{sec-Application}

In this section, we demonstrate the application of dimensional clustering to various datasets.

\subsection{Random walk}

\begin{figure}[htb]
   \begin{center}
    \includegraphics[width=0.8\linewidth]{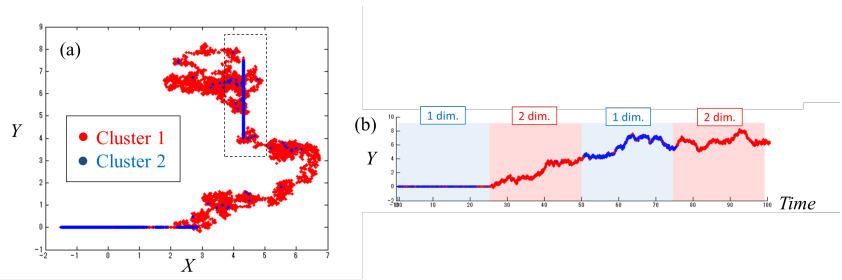}
    \caption{ Dimension-switching random walk \label{fig-RandomWalk}}
    \end{center}
  \end{figure}

The first example is a dimension-changing random walk.
The data set is generated by two distinct, alternating sub-processes. One is a random walk along a line, and the other is a random walk on a plane. Each sub-process is run twice, and the two 1-dimensional phases operate in orthogonal directions. As these processes differ in dimension, our algorithm is expected to detect that the data points from the 1-dimensional process were generated differently than those from the 2-dimensional process. The data analyzed is a series of X-Y coordinates without time. Figure \ref{fig-RandomWalk}(a) shows the dimensional clustering analysis for this data set (two estimated clusters are in different colors).  Figure \ref{fig-RandomWalk} (b) shows the temporal structure underlying the random walk data set. Except for a small number of errors, the algorithm correctly detect two distinct clusters and assigns most of the data points to the correct clusters. This simple case confirms that dimensional clustering correctly detects distinct dimensional components in a data set.
Note that some data points generated by the one-dimensional walk overlap with those of the two-dimensional one (the broken box in Figure \ref{fig-RandomWalk}).
It would be difficult to discriminate them using conventional clustering with this spatial overlap.

\subsection{Image Data}
\begin{figure}[h]
\begin{center}
\includegraphics[width=0.8\linewidth]{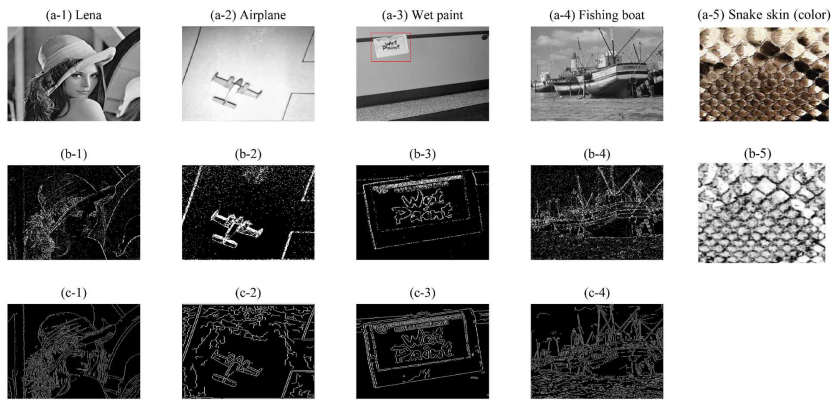}
\caption{(a) Test images, (b) Feature clusters detected by dimensional clustering, and (c) Results of Canny edge detection on the test images.}
\label{fig-improc}
\end{center}
\end{figure}

For the second application of dimensional clustering, we analyzed a several standard test images in the field of image processing. These test images are presented as (a-1) to (a-4) in Figure \ref{fig-improc}. We obtained ``Lena'' and ``Wet paint'' from Mike Wakin's website \cite{Wakin}, 
and ``Airplane'' and ``Fishing boat'' from the USC SIPI Database \cite{USC}.


In such analyses, the first substantial problem one must consider is that of preparing the data for analysis. For each test image in Figure \ref{fig-improc}, the second dimensional clustering algorithm was applied to a 3-dimensional or 5-dimensional data set of vectors listing the $x$-coordinate, $y$- coordinate, and grayscale value or RGB values of each pixel in the images.
The algorithm discovered three or four clusters in each data set. Among these, there was always a ``feature'' cluster which  discerned sharp changes in hue. Images (b-1) to (b-4) of Figure \ref{fig-improc} show the pixels in each image belonging to the corresponding feature clusters. Images (c-1) to (c-4) of Figure \ref{fig-improc} show the results of running the Canny edge detection algorithm \cite{Canny1986} on the same test images, and are provided for comparison.
Figure \ref{fig-improc}(b-3) highlights the region in the feature cluster detected in Figure \ref{fig-improc}(a-3) which contains the sign. This shows that the dimensional clustering algorithm very clearly extracts its primary lettering.

As far as we know, many edge detection methods and other types of filters are readily applicable only to grey scale image (or single-valued other feature). If one wishes to apply the Canny edge detector to a colored image, one would typically need to pick either a basic color (e.g. RGB) or another channel (e.g. hue, satulation, lightness, and value) to which to apply the filter.
Figure \ref{fig-improc} (b-5) shows that dimensional clustering captures the edge-like patterns of a rather complex mixture of color texture.
The fact that dimensional clustering is readily applicable to both grey scale and color images without any substantial preprocesses may inspire new techniques in the field of image processing.

\subsection{Action segmentation}

\begin{figure}[h]
\begin{center}
\includegraphics[width=0.98\linewidth]{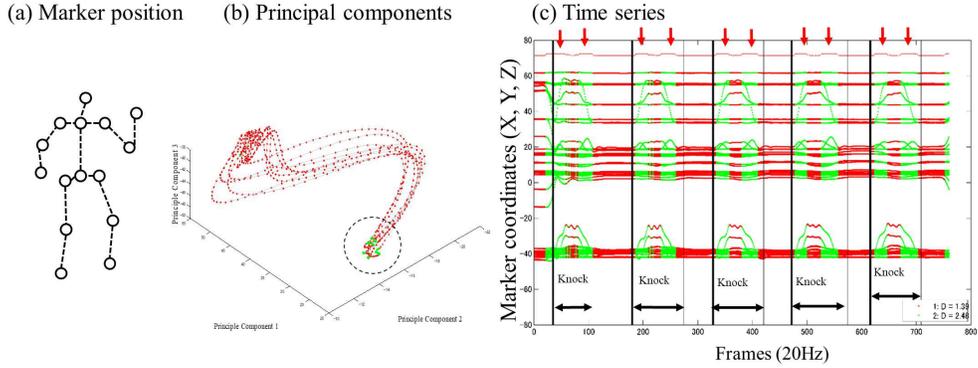}
\caption{Action segmentation task: (a)Positions of 15 bodily markers, (b) the three-dimensional principal component analysis of a knocking trajectory, and (c) time series of a knocking.}
\label{fig-action}
\end{center}
\end{figure}

The third case study is action segmentation. The human body consists of hundreds of joints and muscles controlled by some complex mechanism. Any motion generated by such a system is supposed to reflect the states of one or more dynamical systems.

Here we apply dimensional clustering to human bodily movements, coded as multi-dimensional time series, and try to find some underlying units of actions.
Although it is difficult to identify which body parts and which kind of movements may be involved with a particular kind of action,
any well-performed action is supposed to form a smooth trajectory in the multi-dimensional state space in which we represent the positions of different body parts over time.
As already seen in the random walk case study, dimensional clustering can detect such latent switch in the generating system
without specifically identifying which direction or dimension it is. Thus, we expect that an ``action'' or a ``switch from one action to another'' would be automatically detected as distinct dimensional clusters.

Specifically, we employ the action dataset collected by Ma et al. \cite{Ma2006}, which includes three right-hand action types (knock, throw, and lift) performed by 30 actors.
In this database, each movement is recorded for approximately one minute at 20Hz by 15 bodily markers (each is in (X,Y,Z) coordinate, 45 dimension in total) (Figure \ref{fig-action}(a)), and one of three types of hand actions was performed repeatedly 5 times in each trial. Figure \ref{fig-action} (b) shows the visualization of the first three principal components of the 45 dimensional time series of a knocking trial, and Figure \ref{fig-action} (c) shows the 45 time series of the same knocking data as a function of frames.

By applying dimensional clustering to this 45 dimensional time series of knocking,
we found the two distinct dimensional clusters ($d=1.39, 2.48$) shown in red and green points in these figures.
This motion database includes human-coded action intervals, which are shown by the 5 black arrows in Figure \ref{fig-action} (c).
On the time series plot, the dimensional cluster with higher dimension (green) corresponds with the beginning and end of knocking actions.
This cluster of points in the three-dimensional PCA plot (Figure \ref{fig-action} (b)) captures the non-smooth change (singular) of high-dimensional trajectories.
For the other two action types (throw and lift) and different actors, we typically observed similar results that one of estimated dimensional clusters captures the begining and end of the actions. 

This confirms that dimensional clustering successfully detect the change across actions, from standing to knocking (throwing, lifting) or the other way around.
No particular preprocessing was applied to the data. The entire time series of all markers was submitted to dimensional clustering. Dimensional clustering seems to have captured abstract action semantics invariant across trials, actors, and action types.

\section{Concluding remarks}

The three applications that we presented here show that dimensional clustering is capable of capturing semantically
relevant information in its clusters with no expert supervision and no preprocessing. Note that the semantics
that can be captured by dimensional clustering are limited to those that can be captured purely through dimensional
characteristics. For example, dimensional clustering provides no way of telling apart the two linear phases
in the random walk example -- even though they operated in different directions.

Still, the technique has great potential. We are planning to open it up to the community and are in the process of 
releasing as open source an implementation of the algorithm presented in this paper.





\bibliographystyle{plain}
\bibliography{NNDimReferences}

\end{document}